\documentclass[runningheads]{llncs}

\usepackage[year=2024,ID=1961]{eccv}
\usepackage{eccvabbrv}

\usepackage{graphicx}
\usepackage{booktabs}
\usepackage{bbm}
\usepackage{multirow}
\usepackage{algorithm}
\usepackage{algorithmicx}
\usepackage{bbding}
\usepackage[accsupp]{axessibility}  % Improves PDF readability for those with disabilities.

\usepackage[pagebackref,breaklinks,colorlinks,citecolor=eccvblue]{hyperref}
\usepackage{orcidlink}

\begin{document}

\title{Neural Video Fields Editing} 



% \author{
%     % Author 1\thanks{Equal contribution} \qquad
%     % Author 2\footnotemark[1] \qquad
%     % Author 3 \\
%     % Institute \\
%     % {\tt\small \{email, addresses\}@inst.edu}
%     Shuzhou Yang$^{1,2}$ \qquad 
%     Chong Mou$^{1}$ \qquad
%     Jiwen Yu$^{1}$ \qquad 
%     Yuhan Wang$^{1,2}$\\
%     Xiandong Meng$^{2}$ \qquad 
%     Jian Zhang$^{1}$\Envelope\\[0.9em]
%     $^{1}$SECE, Peking University \qquad $^{2}$Peng Cheng Laboratory
%     \\[0.9em]
%     \url{https://nvedit.github.io/}
% }

\author{Shuzhou Yang\inst{1,2} \qquad Chong Mou\inst{1} \qquad
Jiwen Yu\inst{1} \qquad Yuhan Wang\inst{1,2} \\ Xiandong Meng\inst{2} \qquad Jian Zhang\inst{1}\Envelope}


\institute{$^{1}$SECE, Peking University \qquad $^{2}$Peng Cheng Laboratory}
% \\ \email{szyang@stu.pku.edu.cn}, \email{zhangjian.sz@pku.edu.cn}}

\maketitle

\begin{center}
\captionsetup{type=figure}
\vspace{-1.5em}

\newcommand{\imwidth}{\textwidth}

\begin{tabular}{@{}c@{}}

\parbox{\imwidth}{\includegraphics[width=\imwidth, ]{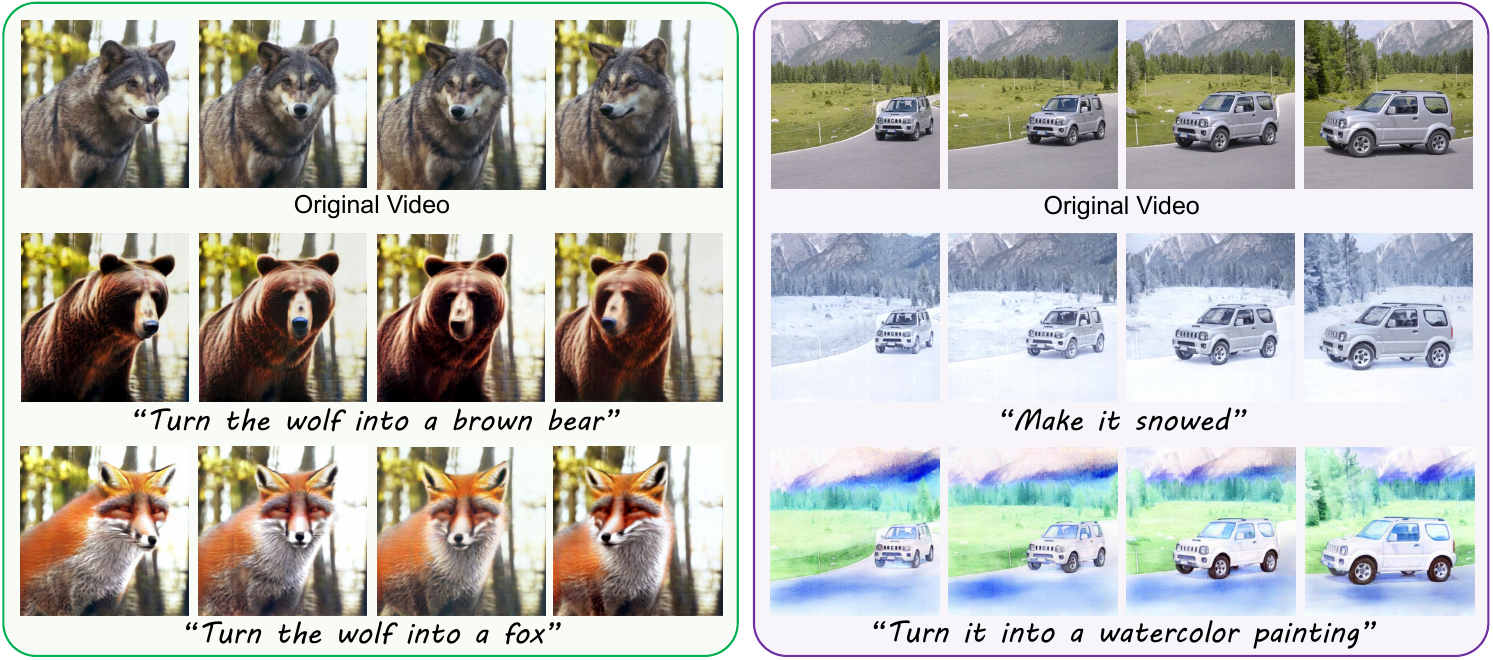}}

% \vspace{1em}
\end{tabular}
\vspace{-0.7em}
    \captionof{figure}{\textbf{NVEdit} enables various editing options, including shape variation, scene change, and style transfer, while preserving original motion and semantic layout. Due to its efficient encoding rate, long videos with hundreds of frames even be edited well.}
    \label{fig:teaser}
    % \vspace{-1.em}
\end{center}

\begin{abstract}
  Diffusion models have revolutionized text-driven video editing. However, applying these methods to real-world editing encounters two significant challenges: (1) the rapid increase in GPU memory demand as the number of frames grows, and (2) the inter-frame inconsistency in edited videos. To this end, we propose \textbf{NVEdit}, a novel text-driven video editing framework designed to mitigate memory overhead and improve consistent editing for real-world long videos. Specifically, we construct a neural video field, powered by tri-plane and sparse grid, to enable encoding long videos with hundreds of frames in a memory-efficient manner. Next, we update the video field through off-the-shelf Text-to-Image (T2I) models to impart text-driven editing effects. A progressive optimization strategy is developed to preserve original temporal priors. Importantly, both the neural video field and T2I model are adaptable and replaceable, thus inspiring future research. Experiments demonstrate the ability of our approach to edit hundreds of frames with impressive inter-frame consistency. Our project is available at: \url{https://nvedit.github.io/}.
\end{abstract}

\section{Introduction}
\label{sec:intro}

Text-driven video editing aims to faithfully edit high-quality videos following the user-given prompts. Considering the advancement of Text-to-Image (T2I) methods \cite{T2I-Adapter, p2p, sdedit}, adopting the pre-trained T2I models to edit videos has attracted wide attention \cite{Esser_2023_ICCV, Dreamix, Wang_2023_arxiv, Chai_2023_ICCV}. However, T2I models inherently lack temporal priors, resulting in inevitable inconsistency across edited frames. Hence, editing videos more temporal-consistently has become a challenging task.

% Due to the limited paired video-prompt data, using videos to train models is expensive \cite{MakeAVideo, instructvid2vid}. Considering the advancement of Text-to-Image (T2I) methods \cite{T2I-Adapter, p2p, sdedit}, adopting the pre-trained T2I models to edit videos has attracted wide attention \cite{Esser_2023_ICCV, Dreamix, Wang_2023_arxiv, Chai_2023_ICCV}. However, T2I models inherently lack cross-frame semantic consistency, resulting in inevitable flickering between edited frames. Hence, editing videos more temporal-consistently has become a challenging task.

In pursuit of this objective, numerous algorithms improve cross-attention operations \cite{Wu_2023_ICCV, Rerender_A_Video} and DDIM \cite{DDIM} inversion \cite{Mokady_2023_CVPR, Ceylan_2023_ICCV, Shin_2023_arxiv, difflle} to ensure superior generation while maintaining inter-frame coherence. However, these methods still exhibit two principal deficiencies. First, although various constraints and attention controls are introduced to the generation process, inter-frame inconsistency still exhibits in the edited videos, such as object deformation and texture variations. That is because T2I models inherently lack temporal motion priors. The second drawback is the rapid increase in GPU memory demands as more frames are edited, while limited hardware devices constrain the length of videos they can edit. Since editing long videos with hundreds of frames is significant for real-world applications, existing T2I models cannot achieve satisfactory effects in videos. Neural representation \cite{nerf, NeRV, enerv} may be a feasible tool to decrease memory overhead while improving inter-frame consistency, as it is naturally a smooth representation \cite{SIREN} and has a high compression rate for videos \cite{NeRV}. Some current works have attempted this proposal \cite{Kasten_2021_TOG, CoDeF} but brought other limitations, such as inaccuracies in complex motion or viewpoint changing, \textit{etc.}

To address these challenges, a memory-efficient framework is developed for consistent text-driven video editing, avoiding intricate attention-related operations or fine-tuning for T2I models. Specifically, our approach comprises two stages: \textbf{video fitting} and \textbf{field editing}. In the first stage, we construct a Neural Video Field (NVF) with tri-plane encoding and MultiLayer Perceptron (MLP) decoding, efficiently modeling temporal and content priors of a given video. Benefiting from its encoding efficiency, even lengthy videos with hundreds of frames can be compactly represented as a signal field. In the second stage, the trained NVF is updated to incorporate text-based editing effects. During each iteration, NVF renders a frame, and a T2I model is employed to edit it based on the provided text. These edited frames serve as pseudo Ground Truths (GTs) to optimize the parameters of NVF. Since NVF has been initially trained on the original video, it retains robust temporal priors after optimization while manifesting the desired editing effects. This strategy of using \textbf{N}eural \textbf{V}ideo fields for video \textbf{Edit}ing is termed \textbf{NVEdit}. Our NVEdit also possesses inherent intriguing properties of neural representation, \textit{e.g.}, frame interpolation, which has been explored in recent works \cite{NeRV, NVP}. It means that NVEdit can predict frames not present in the original video while exhibiting coherent editing effects. In addition, both the NVF and the T2I model within NVEdit are adaptable, allowing for seamless improvements or replacements based on diverse requirements. In this paper, we majorly adopt InstructPix2Pix (IP2P) \cite{InstructPix2Pix} for editing effects. We observed that for local editing, IP2P tends to modify regions that should be preserved inaccurately, impacting video editing outcomes. To address this, an auxiliary mask is designed to constrain the editable region without additional calculation, resulting in Instruct-Pix2Pix+ (IP2P+). This version not only enhances local editing capabilities but also demonstrates the potential for T2I model improvement within NVEdit. Overall, our contributions are as follows:
\begin{itemize}
    \item[$\bullet$]
    We develop NVEdit, using a neural video field to fit long videos and refining it with off-the-shelf T2I models for editing effects. It allows for seamless improvement or replacement of both the T2I models and the video field.
    \item[$\bullet$]
    We propose an auxiliary mask to promote the local editing capability of IP2P, thus benefiting our video editing. It also highlights our potential for T2I model improvement.
    \item[$\bullet$]
    Extensive experiments demonstrate that the proposed method edits long videos with hundreds of frames more consistently while exhibiting promising editing effects.
    \item[$\bullet$]
    NVEdit capitalizes on the unique properties of neural representation, including frame interpolation. The interpolated frames exhibit coherent editing effects seamlessly, without any fine-tuning or post-processing operations.
\end{itemize}

\section{Related Work}
\label{sec:rw}

\subsection{Text-driven Video Editing}
Editing videos through user-given texts has attracted wide attention \cite{VDM, Wang_2024_arxiv}. A straightforward idea is to train a model with video-text pairs. Ho~\textit{et al.} \cite{VDM} first proposed to train a diffusion model from videos and images but required paired video-text data, which is expensive to obtain. To reduce the reliance on video data, Singer~\textit{et al.} \cite{MakeAVideo} learned visual-text correlation from paired image-text data and captured motion information from videos. Qin~\textit{et al.} \cite{instructvid2vid} synthesized video-prompt triplets with off-the-shelf models, but training with video data is expensive. As the advancement of conditional diffusion models \cite{LDM, DALL-E3, SDXL} promotes text-to-image generation \cite{GLIDE, Ruiz_2023_CVPR, Imagen}, exploiting their effective vision priors to edit videos has attracted wide attention. The key challenge is ensuring inter-frame consistency. Wu~\textit{et al.} \cite{Wu_2023_ICCV} fine-tuned a T2I model to fit a specific video for temporal modeling. Khachatryan~\textit{et al.} \cite{Khachatryan_2023_arxiv} simulated motions by translating latent. Qi~\textit{et al.} \cite{FateZero} fused attention features before and after editing for consistency. Wang~\textit{et al.} \cite{vid2vid-zero} developed dense spatial-temporal attention to learn motion. Geyer~\textit{et al.} \cite{tokenflow} only edited key frames and propagated to the whole video. Liew~\textit{et al.} \cite{Liew_2023_arxiv} decoupled content, structure, and motion signals to model coherent features. However, these methods require rapid increased memory to edit more frames, limiting the potential for editing long videos.

To this end, NVEdit encodes videos in a memory-efficient manner to achieve long video editing, which has more significant value in practical application.

\subsection{Neural Representation for Videos}
Using coordinate-based neural networks to represent visual signals is an emerging practical technology \cite{nerfocus, Yang_2023_ICCV, NVP, Cheng_2023_CVPR, Cheng_2023_arxiv}. Sitzmann~\textit{et al.} \cite{SIREN} first encoded a video by a network, whose input is the spatio-temporal coordinate ($\it{x,y,t}$) and output is a pixel value. But its reconstruction quality is limited and training takes too long. To this end, Chen~\textit{et al.} \cite{NeRV} replaced input with the time stamp $\it{t}$ to output the $\it{t}$-th frame of the target video for better performance. Li~\textit{et al.} \cite{enerv} disentangled spatio-temporal context to accelerate mapping from the input coordinates to the output frames. Bai~\textit{et al.} \cite{psnerv} chose to divide a frame into several patches, which reduces training difficulty and improves fitting accuracy. Kim~\textit{et al.} \cite{NVP} developed an explicit-implicit hybrid structure to improve fitting speed and accuracy greatly.

Recently, some attempts at video editing appeared. Kasten~\textit{et al.} \cite{Kasten_2021_TOG} utilized MLPs to map video pixels to Atlas maps, and edited videos by operating on Atlas maps. Further, Ouyang~\textit{et al.} \cite{CoDeF} mapped video content to a canonical image and edited it to realize video editing. However, the strategy of extracting video content to a representation medium and editing it naturally has two disadvantages. First, for videos with significant content changes (such as large movements or long range motion), it is hard to map all content to a single medium accurately. Second, their editing effect entirely relies on the image editing of the medium. But T2I models may perform poorly in certain cases, \textit{i.e.}, lacking robustness.

In this paper, we optimize a well-designed neural video field for editing. It not only maintains temporal consistency but is also robust to T2I model failures in certain situations.

\section{Method}
Our goal is to edit videos based on user-given instructions. In \cref{subsec:pre}, we first introduce some preliminaries, including latent diffusion models, a well-known T2I tool (Instruct-Pix2Pix \cite{InstructPix2Pix}), and the principles of the neural video field. Then, we discuss the details of NVEdit in \cref{subsec:vf}. It attains temporal priors in the fitting stage and imparts editing effects with a neural video field. Finally, in \cref{subsec:ip2p}, we showcase IP2P+, an enhanced version of IP2P, to improve editing quality. In this paper, we majorly adopt IP2P+ for editing effects. However, it is worth noting that both the video field and T2I model are replaceable and improvable.

\subsection{Preliminaries}
\label{subsec:pre}
\textbf{Latent Diffusion Models (LDMs).} LDMs \cite{LDM} encode the input image to a latent embedding $\textbf{z}_0$, and receive a prompt $\emph{P}$ as the condition, which is also encoded to an embedding $\textbf{c}_P$. Following DDPM \cite{DDPM}, according to the step $\it{t}$, ${\textbf z}_0$ is diffused to $\textbf{z}_t$ through random noise $\boldsymbol{\varepsilon}$. LDMs aim to predict the added noise by minimizing the following objective:
\begin{eqnarray}
\begin{aligned}
    \label{shi1}
    \underset{\boldsymbol{\theta}}{\min} \, \mathbbm{E}_{\textbf{z}_0, \boldsymbol{\varepsilon} \sim \mathcal{N}(0,I), t} \|\boldsymbol{\varepsilon} - \varepsilon_{\boldsymbol{\theta}}(\textbf{z}_t, t, \textbf{c}_P)\|_2^2.
\end{aligned}
\end{eqnarray}
During inference, given a random latent $\textbf{z}_T$ and a prompt $\emph{P}$, the trained model predicts noise $\varepsilon_{\boldsymbol{\theta}}(\cdot)$ for $\it{T}$ steps to generate an image, which meets the textual condition $\emph{P}$. This accomplishment has prompted further exploration in image generation \cite{controlnet}, image editing \cite{InstructPix2Pix}, video generation \cite{AnimateDiff}, and protection \cite{Zhang_2023_arxiv}.

\textbf{Instruct-Pix2Pix.} Based on LDMs which incorporate textual condition $\textbf{c}_P$, IP2P \cite{InstructPix2Pix} introduces an image condition $\textbf{c}_I$ to generate images that exhibit similar semantic layouts to $\textbf{c}_I$ while adhering to $\textbf{c}_P$. Given a random embedding $\textbf{z}_T$, IP2P predicts noise $\varepsilon_{\boldsymbol{\theta}}(\cdot)$ for $\it{T}$ steps to get results. At each step $\it{t}$, its predicted noise can be expressed as:
\begin{eqnarray}
\begin{aligned}
    \label{shi2}
    \varepsilon_{\boldsymbol{\theta}}(\textbf{z}_t, \textbf{c}_I, \textbf{c}_P) =& \varepsilon_{\boldsymbol{\theta}}(\textbf{z}_t, \varnothing, \varnothing) \\ &+ s_I \cdot (\varepsilon_{\boldsymbol{\theta}}(\textbf{z}_t, \textbf{c}_I, \varnothing) - \varepsilon_{\boldsymbol{\theta}}(\textbf{z}_t, \varnothing, \varnothing)) \\ &+ s_P \cdot (\varepsilon_{\boldsymbol{\theta}}(\textbf{z}_t, \textbf{c}_I, \textbf{c}_P) - \varepsilon_{\boldsymbol{\theta}}(\textbf{z}_t, \textbf{c}_I, \varnothing)),
\end{aligned}
\end{eqnarray}
where $s_I$ and $s_P$ are the weights for image and text conditions respectively. The first term means random sampling, the second term guides sampling based on the image condition, while the third term introduces the textual condition. Compared to previous text-driven methods \cite{p2p, diffedit}, which mainly belong to the prompt-to-prompt type and need well-designed paired texts, IP2P enables users to edit images more conveniently, \textit{i.e.}, directly using one oral instruction to edit. Existing method \cite{instructnerf2nerf} has exploited the prior of IP2P to make 3D editing easier. However, there is still no research on directly benefiting from the off-the-shelf IP2P to conveniently edit videos with oral instructions.

\textbf{Neural video fields.} Recent works propose to represent video information through neural networks \cite{Zhang_2021_arxiv, NeRV, NVP}, which can mainly be simplified to a mapping function as:
\begin{eqnarray}
\begin{aligned}
    \label{shi3}
    \widetilde{\textbf{V}}[x, y, t] = f_{\boldsymbol{\theta}}(x, y, t),
\end{aligned}
\end{eqnarray}
where $\it{t}$ is the temporal stamp of a frame in the video, $\it{x}$ and $\it{y}$ are the horizontal and vertical coordinates of a pixel on the frame. $\widetilde{\textbf{V}}$ is the target video signal. Their training purpose is to update parameters $\boldsymbol{\theta}$ to fit the given video. During inference, given the coordinates of all pixels from a video, the model $f_{\boldsymbol{\theta}}(\cdot)$ will output the entire video.

\subsection{NVEdit}
\label{subsec:vf}
In this section, we discuss how to build a Neural Video Field (NVF) to achieve editing in a memory-efficient manner. We first introduce the structure of NVF, then showcase the process of video fitting, field editing, and inference.

\begin{figure*}[t]
  \centering
  \begin{subfigure}{1\linewidth}
    \includegraphics[width=1.\linewidth]{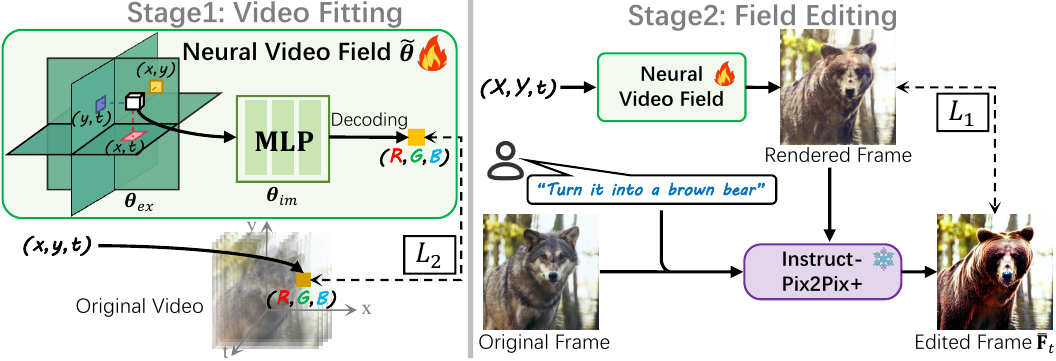}
  \end{subfigure}
  \vspace{-2.em}
  
  \caption{Workflow of our NVEdit. It contains two stages: video fitting and field editing. As shown in the left part, we first train a neural video field to fit a given video for temporal priors. Then, on the right, the rendered frame is edited and used to optimize the trained field to impart editing effects. As the video field has learned temporal priors, the optimized field consistently renders a video with edited content.}
  \label{totnet}
  \vspace{-1.5em}
  
\end{figure*}

\textbf{Model structure.} As previous works \cite{NVP, eg3d} have validated that the explicit-implicit hybrid structures benefit fitting speed and accuracy. We thus build a hybrid model to represent video signals. As shown in the left part of \cref{totnet}, our explicit module contains tri-plane and sparse grids. The former is three feature maps designed to record video features and the latter is a set of parameter grids sparsely distributed across the $x-y-t$ coordinate system of the video. Both of them are trainable parameters, denoted as $\boldsymbol{\theta}_{ex}$. Given a spatio-temporal coordinate $(x, y, t)$, we extract features from its neighbor grids and tri-plane, represented as the white cube in \cref{totnet}. Our implicit part is an MLP that decodes the aggregated features into a pixel value, whose parameters are $\boldsymbol{\theta}_{im}$. We update $\widetilde{\boldsymbol{\theta}} = \left \{ \boldsymbol{\theta}_{im}, \boldsymbol{\theta}_{ex} \right \}$ to fit a video $\textbf{V}$ and then achieve editing effects.

\textbf{Video fitting stage.} In each iteration, we randomly sample a batch of pixels from $\textbf{V}$, denoted as $\textbf{V}[\boldsymbol{\psi}]$, and their spatio-temporal coordinates set $\boldsymbol{\psi}$. $\textbf{V}[\boldsymbol{\psi}]$ are considered as the Ground Truths (GTs) and $\boldsymbol{\psi}$ serves as the input for NVF, based on which NVF predicts pixel values. We enforce its output to be similar to GT, expressed as:
\begin{eqnarray}
\begin{aligned}
    \label{shi4}
    \mathcal{L}_{\rm{fit}}(\widetilde{\boldsymbol{\theta}}) = \underset{(x,y,t)\in\boldsymbol{\psi}}{\sum} \|\textbf{V}[x,y,t] - f_{\widetilde{\boldsymbol{\theta}}}(x, y, t)\|_2^2.
\end{aligned}
\end{eqnarray}
As the batch size is independent of the video length, our GPU memory demand remains constant. Notably, we supervise training with randomly sampled pixels across the entire video rather than a batch of frames, since we found this scheme yields faster convergence and slightly better fitting accuracy compared to training frame-by-frame. In this stage, NVF learns the temporal and motion priors of the given video, which facilitates subsequent consistent editing.

\textbf{Field editing stage.} We update the parameters of the trained NVF with a T2I model to impart editing effects, as shown in the right of \cref{totnet}. In each iteration, we input a coordinate set ($\textbf{\emph{X}}, \textbf{\emph{Y}}, \it{t}$), where $\it{t}$ means the frame number, $\textbf{\emph{X}}$ and $\textbf{\emph{Y}}$ are the sets of horizontal and vertical coordinates of pixels respectively. These three elements represent the coordinates of all pixels in a frame. NVF renders a frame based on them, as depicted in \cref{totnet}. Subsequently, the original frame is retrieved from the input video based on $\it{t}$ as the image condition, whilst the user-provided instruction serves as the text condition. Consequently, we employ a T2I model to process the rendered frame and get an edited version $\overline{\textbf{F}}_t$, which is used to optimize NVF. The objective function is: \begin{eqnarray}
\begin{aligned}
    \label{shi5}
    \mathcal{L}_{\rm{opt}}(\widetilde{\boldsymbol{\theta}}) = \|f_{\widetilde{\boldsymbol{\theta}}}(\textbf{\emph{X}}, \textbf{\emph{Y}}, t) -\overline{\textbf{F}}_t\|_1.
\end{aligned}
\end{eqnarray}
Note that we only update NVF and fix the T2I model. Different from \cref{subsec:vf} that samples pixels randomly, here we supervise NVF frame-by-frame because T2I models exhibit optimal performance on complete images rather than irregularly sampled pixels. As only one frame is edited per iteration, our GPU memory requirement is constant and independent of the video length. Importantly, we adopt IP2P \cite{InstructPix2Pix} as the T2I model, and further improve it for better editing effects, which is discussed in \cref{subsec:ip2p}. It also proves that the T2I model of NVEdit is \textbf{replaceable} and \textbf{improvable}, related further experiments are given in \cref{sec:4.4}.

In addition, a progressive optimization strategy is developed for temporal coherence. Specifically, At the beginning of the field editing stage, we assign a lower weight to the instruction condition, \textit{i.e.}, setting $s_P$ in Eq.~\ref{shi2} to a small number, and gradually increasing it as the optimization proceeds. This progressive strategy systematically guides the rendered frames toward the desired editing effects step-by-step, while diminishing the updating intensity to NVF at each iteration. Thus, we effectively preserve the temporal priors learned at the video fitting stage. We evaluate the contribution of this strategy in \cref{sec:4.4}.

\textbf{Inference.} After the completion of the aforementioned two stages, we render a coherent video $\widetilde{\textbf{V}}$ with the desired editing effects, using the trained NVF. We provide all pixel coordinates of the entire video, formulated as:
\begin{eqnarray}
\begin{aligned}
\label{shi6}
    \widetilde{\textbf{V}} = f_{\widetilde{\boldsymbol{\theta}}}(\textbf{\emph{X}}, \textbf{\emph{Y}}, \textbf{\emph{T}}),
\end{aligned}
\end{eqnarray}
where $\textbf{\emph{T}}$ is the set of all coordinates on the temporal axis. Results in \cref{sec:Exp} prove that these final rendered videos perform impressive temporal consistency. We believe that is implicitly achieved by two factors: i) In video fitting stage, NVF has learned strong temporal priors from the original video, which is inherently temporal consistent. ii) The optimization strategy in field editing stage preserves the learned temporal priors as it updates model from weak to strong smoothly.

\subsection{Instruct-Pix2Pix+}
\label{subsec:ip2p}
\begin{figure}[t]
  \centering
  \begin{subfigure}{0.9\linewidth}
    \includegraphics[width=1.\linewidth]{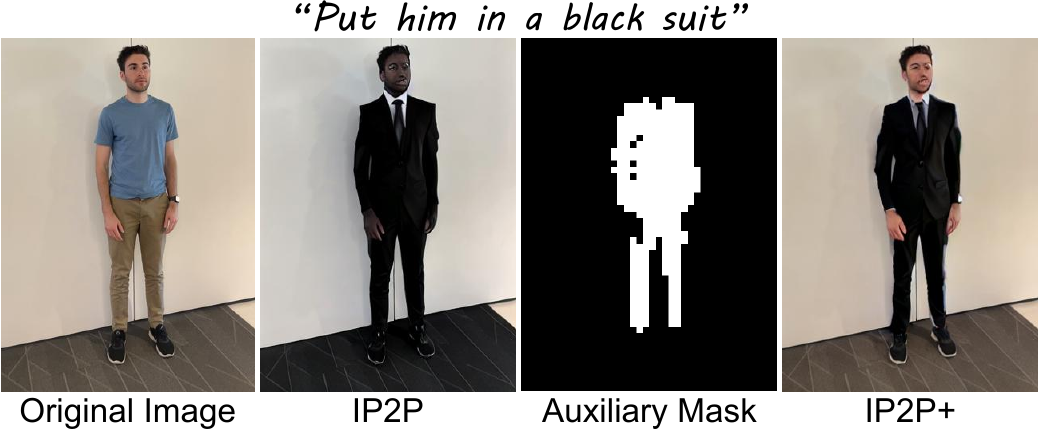}
  \end{subfigure}
  \vspace{-1.em}
  
  \caption{Visualization of the original image, the results of IP2P, the proposed auxiliary mask, and our edited results (IP2P+).}
  \label{mask}
  \vspace{-1.5em}
  
\end{figure}

Although IP2P achieves impressive image editing, for local editing, it alters the region to be retained, as shown in \cref{mask}. We suppose this is an inherent flaw of IP2P as it samples based on Eq.~\ref{shi2}, which only ensures results to be roughly consistent with the image condition, but lacks strong constraints on unedited region. Typically, users prefer instructing specific regions while leaving others untouched. Previous work \cite{diffedit} has used mask to restrict editable regions, but its three-step process requires redundant calculations and well-designed paired texts. To this end, we develop Instruct-Pix2Pix+ to enhance the content fidelity of edited images to the conditional image $I$, without any additional computation.

\textbf{Auxiliary mask.} The predicted noise of IP2P follows Eq.~\ref{shi2}, where the third term indicates the guidance of instructions, as follows:
\begin{eqnarray}
\begin{aligned}
    \label{shi7}
    \boldsymbol{\varepsilon}_{{\boldsymbol{\theta}}_P} = \varepsilon_{\boldsymbol{\theta}}(\textbf{z}_t, \textbf{c}_I, \textbf{c}_P) - \varepsilon_{\boldsymbol{\theta}}(\textbf{z}_t, \textbf{c}_I, \varnothing).
\end{aligned}
\end{eqnarray}
We observed that $\boldsymbol{\varepsilon}_{{\boldsymbol{\theta}}_P}$ represents the relationship between the generated image and text condition. Hence, a simple yet effective auxiliary mask is proposed. We apply normalization and dimensionality reduction to $\boldsymbol{\varepsilon}_{{\boldsymbol{\theta}}_P}$, getting a gray-scale image $\textbf{G}_{x,I,T}$, in which a higher value $x$ means the pixel is more likely to be relevant to instruction, and we allow it to change. In contrast, a lower $x$ signals that the pixel is not require change. A mask threshold $\tau \in [0,1]$ is set on $\textbf{G}_{x,I,T}$ to get our auxiliary mask, $\textbf{M}_{x,I,T} = \mathbbm{1}(\textbf{G}_{x,I,T} \geq \tau)$. As $\textbf{M}_{x,I,T}$ is derived from $\boldsymbol{\varepsilon}_{{\boldsymbol{\theta}}_P}$, a sub-term of the predicted noise in generation process. $\textbf{M}_{x,I,T}$ is regarded as an auxiliary product obtained during editing and termed as `auxiliary mask'.

% Notice that we only use $\boldsymbol{\varepsilon}_{{\boldsymbol{\theta}}_P}$, which is an essential element in generation process, to get $\textbf{M}_{x,I,T}$, without any additional calculation. 

\textbf{Generation.} According to \cite{DDIM}, IP2P samples a latent $\textbf{z}_{t-1}$ at step $\it{t-}$1. With our auxiliary mask $\textbf{M}_{x,I,T}$, the sampling latent of IP2P at timestep $\it{t-}$1 can be tuned as:
\begin{eqnarray}
\begin{aligned}
    \label{shi8}
    \textbf{z}_{t-1} = \textbf{z}_{t-1} \odot \textbf{M}_{x,I,T} + \widetilde{\textbf{z}}_{t-1} \odot (\mathbf{1}-\textbf{M}_{x,I,T}),
\end{aligned}
\end{eqnarray}
where $\textbf{z}_{t-1}$ is the latent generated by IP2P and $\widetilde{\textbf{z}}_{t-1}$ is the noisy latent of the conditional image $I$, which is also added the same level of noise as $\textbf{z}_{t-1}$ according to \cite{DDIM}. We thus only modify the latent of the most relevant region while preserving the original latent of other regions. Visual results are provided in \cref{mask} for a more intuitive illustration. Obviously, the vanilla IP2P inaccurately alters regions that should be preserved, such as the floor. With $\textbf{M}_{x,I,T}$, IP2P+ edits the target objects and preserves original information well. Note that we only use auxiliary mask in the case of local editing. This enhanced version of IP2P also validates that the T2I model of NVEdit is adaptable.

\section{Experiments}
\label{sec:Exp}

\subsection{Implementation Details}
To generate edited frames for optimizing NVF, we use the pre-trained IP2P \cite{InstructPix2Pix} as the base model. The proposed auxiliary mask is introduced to improve its local editing capability. Following previous works \cite{NVP, instructnerf2nerf, FateZero}, UVG-HD \cite{UVG}, DAVIS \cite{DAVIS}, and some other in-the-wild videos are adopted to evaluate the editing effect. 

\textbf{Metrics.}
Following \cite{Huang_2023_arxiv, FateZero}, we conduct quantitative evaluations using both objective (with the pre-trained CLIP \cite{CLIP} model) and subjective (with users) indicators. Specifically, three objective metrics are considered: \textbf{i) `Tem-Con'} measures the inter-frame temporal consistency. We only adopt the image encoder from CLIP, and compute the cosine similarity between all pairs of consecutive frames to obtain it. \textbf{ii) `Frame-Acc'} denotes the frame-wise editing accuracy, representing the percentage of frames from the edited video that have a higher CLIP similarity to the target prompt than the source prompt. \textbf{iii) `Vid-Score'} is the average value of the cosine similarity between edited frames and the target prompt, representing semantic disparity. Additionally, we enlist the evaluation of 21 volunteers to compute three subjective metrics, including \textbf{`Edit'}, \textbf{`Image'}, and \textbf{`Temporal'}. These metrics gauge the coherency between edited frames and the target prompt, the quality of edited frames, and the temporal consistency of the edited video, respectively.

\begin{table}[tb]
  \caption{Quantitative evaluation of the edited results obtained from different SOTA approaches, including IP2P \cite{InstructPix2Pix}, Video-P2P \cite{videop2p}, V2V-zero \cite{vid2vid-zero}, TokenFlow \cite{tokenflow}, Fatezero \cite{FateZero}, and CoDeF \cite{CoDeF}. `Tem-Con', `Frame-Acc', and `Frame-Score' are the objective metrics calculated by the pre-trained CLIP \cite{CLIP} model, while `Edit', `Image', and `Temporal' are the subjective metrics scored by users. The best and the second-best scores are highlighted in \textbf{bold} and \underline{underline} respectively.}
  \vspace{-1.em}
  
  \label{compare}
  \centering
      \resizebox{0.95\linewidth}{!}{
    \begin{tabular}{c c|c c c c c c c}
    \toprule[0.5pt]
    \multicolumn{2}{c|}{Metrics} & IP2P & Video-P2P & V2V-zero & TokenFlow & Fatezero & CoDeF & Ours \\
    % & & \cite{InstructPix2Pix} & \cite{videop2p} & \cite{vid2vid-zero} & \cite{tokenflow} & \cite{FateZero} & \cite{CoDeF} & \\
    \bottomrule[0.pt]
    \toprule[0.5pt]
    \multirow{3}{*}{CLIP $\uparrow$} & Tem-Con & 0.959 & 0.867 & 0.853 & 0.939 & 0.956 & \underline{0.972} & \textbf{0.978} \\
	& Frame-Acc & \underline{0.957} & 0.893 & 0.800 & 0.852 & 0.838 & 0.955 & \textbf{0.961} \\
	& Vid-Score & 0.342 & 0.297 & 0.282 & 0.274 & 0.266 & \textbf{0.348} & \underline{0.347} \\ \hline
    \multirow{3}{*}{User $\uparrow$} & Edit & 3.767 & 3.537 & 3.133 & 3.063 & 2.567 & \underline{3.833} & \textbf{4.020} \\
	& Image & \textbf{3.870} & 3.346 & 3.249 & 2.920 & 2.877 & 3.667 & \underline{3.755} \\ 
        & Temporal & 3.267 & 3.238 & 2.067 & 2.992 & 3.183 & \underline{3.967} & \textbf{4.079} \\
   \bottomrule[0.5pt]
    \end{tabular}
	}
 \vspace{-1.em}
\end{table}

\subsection{Comparison with the state-of-the-art}
\label{sec:4.2}
We prove our superior effectiveness by comparing with other five State-Of-The-Art (SOTA) text-driven video editing approaches, including Video-P2P \cite{videop2p}, Vid2Vid-zero (V2V-zero) \cite{vid2vid-zero}, TokenFlow \cite{tokenflow}, Fatezero \cite{FateZero}, and CoDeF \cite{CoDeF}. Furthermore, to demonstrate that the proposed method generates more temporal consistent results than directly applying the T2I model on videos, we fix the random seed of IP2P \cite{InstructPix2Pix} and showcase its results as a baseline. All of these methods are recently proposed. Note that for prompt-to-prompt approaches, to ensure a fair comparison, we elaborate the source-target prompt pairs for them.

% \begin{table}[t]
% 	\centering
% 	\resizebox{\linewidth}{!}{
% 	\begin{tabular}{c|c c c|c c c}
% 	\toprule[1.pt]
% 	\multirow{2}{*}{Method} & \multicolumn{3}{c|}{CLIP Metrics $\uparrow$} & \multicolumn{3}{c}{User Study $\uparrow$} \\ % \cmidrule(r){2-4}  \cmidrule(r){5-7}
%     \cline{2-7}
%         & TCon & FAcc & VScore & Edit & Image & Temp \\ 
%     \bottomrule[0.pt]
%     \toprule
%     Instruct-Pix2Pix &  &  &  &  &  &  \\
%     Video-P2P &  &  &  &  &  &  \\
%     Vid2Vid-zero &  &  &  &  &  &  \\
%     TokenFlow &  &  &  &  &  &  \\
%     Fatezero &  &  &  &  &  &  \\ 
%     CoDeF &  &  &  &  &  &  \\
%     Ours &  &  &  &  &  &  \\
%    \bottomrule[1.pt]
% 	\end{tabular}
% 	}
% 	\vspace{-1.em}
	
% 	\caption{\label{compare} Quantitative evaluation on the enhanced results obtained from different settings. The best and the second best results are highlighted in \textbf{\color{red}red} and \textbf{\color{blue}blue} respectively.}
% 	\vspace{-1.em}
	
% \end{table}

\textbf{Quantitative comparison.}
We present both the objective and subjective metrics of our method and other recent SOTA algorithms. Following \cite{Text2LIVE, FateZero}, for subjective evaluation, we also prepared 16 sets of edited results, shuffled the order of methods within each set, and asked users to rate different results in each set from 1 (the worst) to 5 (the best). As shown in \cref{compare}, the proposed method achieves the best temporal consistency and frame-wise editing accuracy, and demonstrates comparable semantic alignment to the canonical image-based method \cite{CoDeF} in the feature space of the CLIP model. As for user studies, on average, our approach obtains the best user preferences across three aspects.

\begin{figure*}[t]
  \centering
  \begin{subfigure}{1\linewidth}
    \includegraphics[width=1.\linewidth]{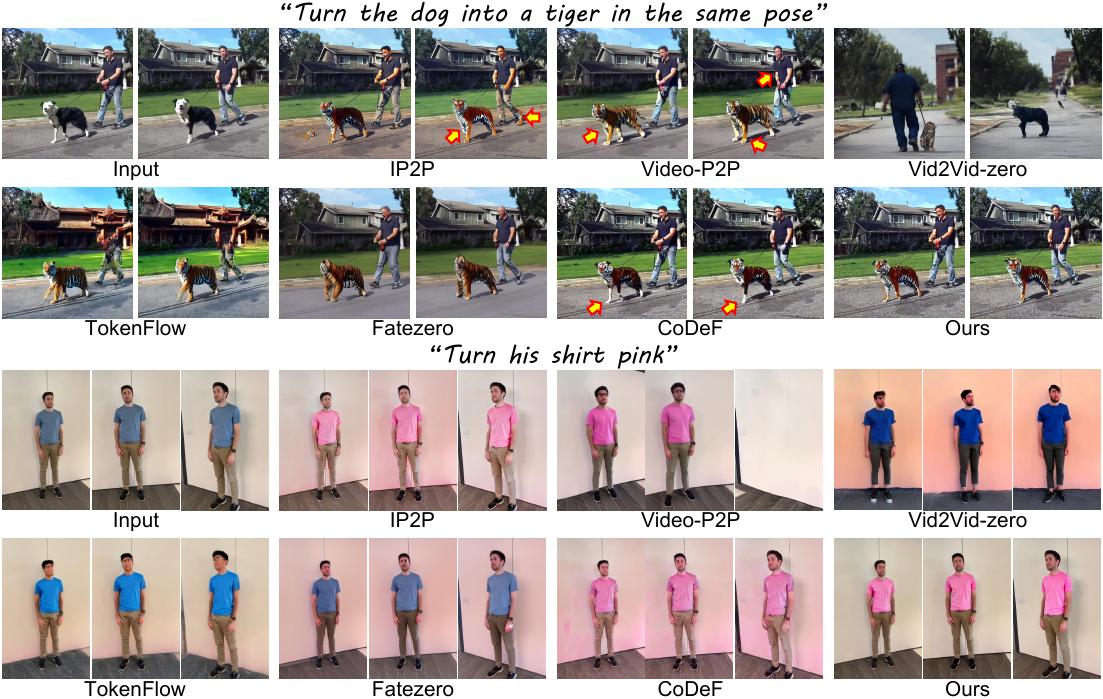}
  \end{subfigure}
  \vspace{-2.em}
  
  \caption{Visual comparison between our method and other SOTA approaches. One can see that IP2P fails to output consistent results, such as the differences in the regions pointed by arrows. Other methods either distort the shape, mistake editing regions, or fail to respond to varying viewpoints. Our approach not only generates temporally coherent content but also controls the area to be edited precisely.}
  \label{compare_visual}
  \vspace{-2.em}
\end{figure*}

\begin{figure}[t]
  \centering
  \begin{subfigure}{0.88\linewidth}
    \includegraphics[width=1.\linewidth]{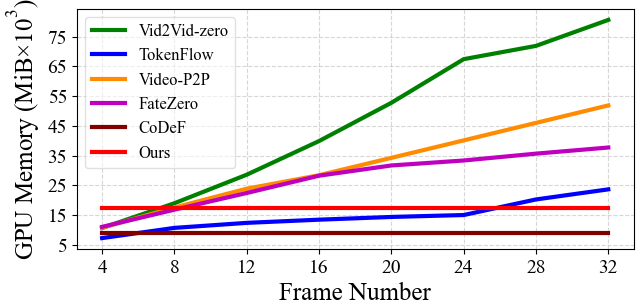}
  \end{subfigure}
  \vspace{-0.5em}
  
  \caption{Comparison of GPU memory overhead. As frames increase, our method only adds minimal memory overhead, which allows it to edit long videos. Although CoDeF \cite{CoDeF} consumes less GPU memory, NVEdit achieves better performance as illustrated in \cref{compare} and it is still memory-efficient compared to most existing methods.}
  \label{memory}
  \vspace{-1.5em}
\end{figure}

\textbf{Qualitative comparison.} In order to provide a more intuitive comparison, we present the visual results of NVEdit and other baselines in two different scenarios in \cref{compare_visual}, showcasing two types of motion changes: the upper scenario changes action, while the lower one changes viewpoint. Instructions used in IP2P and our method are given above. For the prompt-to-prompt methods, we designed original prompts as `\emph{A man is walking his dog on the road in front of the courtyard}' and `\emph{A person stands before the wall}', whilst the target prompts are `\emph{A man is walking his tiger on the road in front of the courtyard}' and `\emph{A person in pink shirt stands before the wall}'. Observing the results, the edited results of frame-wise IP2P vary a lot among different frames (as indicated by the arrows) and fail to preserve unedited region (as shown in the bottom scene). Although other methods develop various techniques, such as attention-related control or null-text inversion, their results incorrectly modify the background and posture, or fail in certain viewpoints. Notice that CoDeF maps video content to a canonical image and edits it through ControlNet \cite{controlnet}. For a fair comparison, we use IP2P to edit the canonical image of CoDeF in this paper, omitting the proposed auxiliary mask as it is a novel tool invented in this work. In the top-row example, the editing of the canonical image is occasionally subpar. However, CoDeF relies entirely on this edit, leading to an unsatisfactory appearance in all of its predicted frames. In contrast, our method, with progressive optimization using numerous pseudo-GTs, accommodates errors in certain pseudo-GTs. Besides, in the lower row, IP2P incorrectly colors the wall, affecting the entire video of CoDeF, whereas our method precisely controls editable region due to IP2P+.

\textbf{GPU memory comparison.} To demonstrate the superior GPU memory efficiency of our method, we use different algorithms to edit frames with a resolution of 480 $\times$ 480 and record their GPU memory overhead. Experiments are conducted on a single NVIDIA A800 GPU. As shown in \cref{memory}, we compare different methods within 32 frames to illustrate their GPU memory growth trends. Except CoDeF \cite{CoDeF} and our method, other approaches require inputting more frames to diffusion models for editing, resulting in a rapid increase in GPU memory. In contrast, by leveraging the efficient encoding of neural representation, both CoDeF and our method add only minor additional memory. Since CoDeF employs an implicit representation, whilst we develop an explicit-implicit hybrid structure as described in \cref{subsec:vf}, our GPU memory overhead is higher than that of CoDeF. However, as noted in \cite{eg3d}, hybrid model enables better representation effect than the implicit one. Although CoDeF maps video content to a canonical image, encoding only motion information, which reduces GPU memory demand. For videos with significant variations, its mapping and encoding tend to be inaccurate. While our method encodes content and motion together through an effective hybrid structure to address it. Comparative experiments also prove our SOTA performance. Our method only needs more GPU memory than CoDeF and is still memory-efficient compared to most existing methods.

\begin{figure*}[t]
  \centering
  \begin{subfigure}{1\linewidth}
    \includegraphics[width=1.\linewidth]{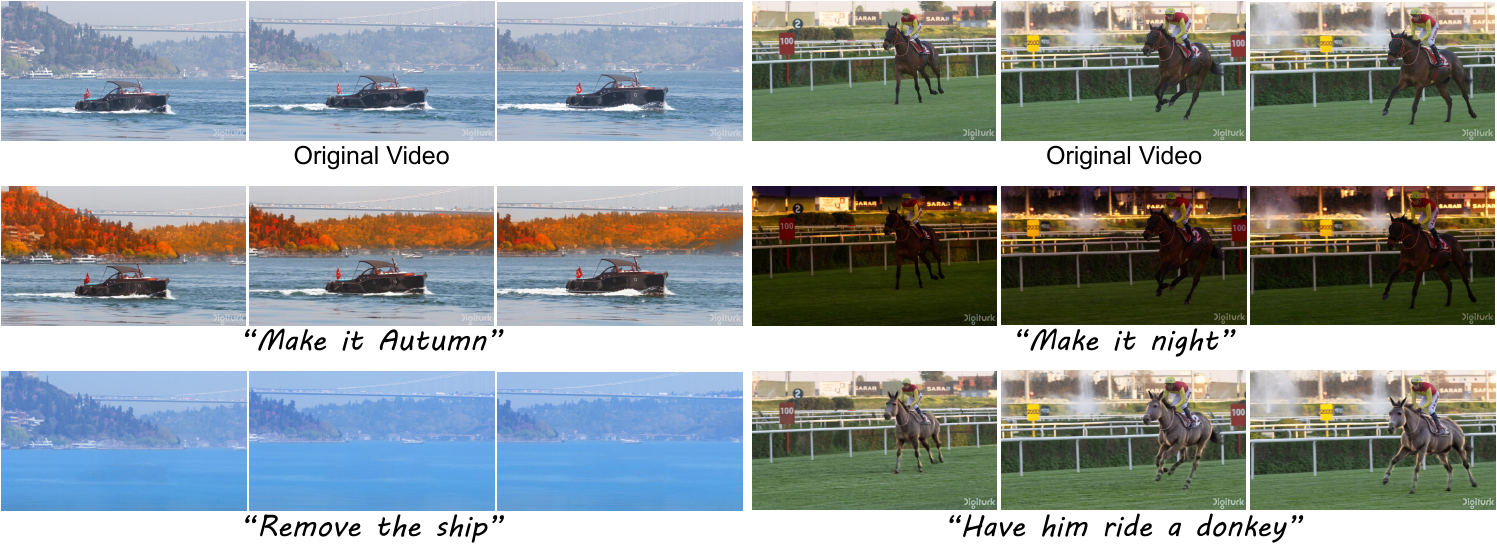}
  \end{subfigure}
  \vspace{-2.3em}
  
  \caption{NVEdit enables temporal consistent, high-quality edits of real-world long videos with hundreds of frames. The edited results faithfully conform to the instruction, while strictly inheriting original information, such as the posture, background, and motion.}
  \label{fig:visual}
  \vspace{-0.5em}
\end{figure*}

\subsection{Application}
Here we demonstrate some other practical applications of NVEdit to establish that it not only enables multiple editing types (\textit{e.g.}, changing shape, editing scene, and style transfer, \textit{etc.}), but also accomplishes frame interpolation without any fine-tuning or post-processing operations.

\textbf{Consistent video editing.} \cref{fig:teaser} has displayed our results on various scenes and instructions, illustrating the temporal consistency and adherence to instructions achieved by our edits. On the left, we replace the wolf with other animals. The added animals maintain the same posture and motion as the wolf. Crucially, the background remains unchanged, whilst inter-frame variation is smooth and consistent throughout the edited video. On the right, we showcase the effects of scene and style editing. Our editing strictly follows the instructions, generating temporal consistent videos while accurately inheriting original features. In addition, due to the impressive encoding efficiency, NVEdit enables long video editing. We present our results on long videos comprising 600 frames in \cref{fig:visual}, whose resolution is 960 $\times$ 544. Again, different types of editing are applied to these videos. One can see that our method faithfully edits frames according to the user-given instructions, while preserving information that does not require modification, such as motion and semantic layout, \textit{etc.} Inter-frame consistency is also well-maintained after the field editing stage, allowing NVEdit to render editing effects without unnatural flickers. We present video results in the supplementary material for a more intuitive illustration.

\begin{table}[t]
    \centering
    % \raggedleft
    \begin{minipage}{.46\linewidth}
    \caption{\label{tab:quan} The quantitative study on other applications of NVEdit.}
    \vspace{-1.2em}
    \centering
    \resizebox{\linewidth}{!}{
    \begin{tabular}{cccc}
    % \toprule[0.5pt]
    \toprule
    Metrics & PSNR $\uparrow$ & SSIM $\uparrow$ & LPIPS $\downarrow$ \\ \midrule
    % Fitting & 39.58 & 0.989 & 0.113  \\
    Interpolation & 33.81 & 0.961 & 0.302  \\
    SR ($\times$4) & 28.95 & 0.812 & 0.246 \\   \bottomrule
    \end{tabular}
    }
    \end{minipage}%
    % \quad
    \hspace{0.2em}
    % \raggedright
    \begin{minipage}{.52\linewidth}
    \caption{\label{tab:abla} Ablation study on progressive optimization. IP2P+ serves as a control group.}
    \vspace{-1.2em}
    \centering
    \resizebox{\linewidth}{!}{
    \begin{tabular}{cccc}
    % \toprule[0.5pt]
    \toprule
    Metrics & Tem-Con $\uparrow$ & Fram-Acc $\uparrow$ & Vid-Score $\uparrow$ \\ \midrule
    IP2P+ & 0.966 & 0.957 & 0.309 \\
    Fix $s_P$ & 0.959 & 0.960 & 0.313 \\
    % NVEdit &  0.978 & 0.961 & 0.347 \\
    %\bottomrule[0.5pt]
    \bottomrule
    \end{tabular}
    }
	
    \end{minipage}
\vspace{-1.1em}
\end{table}

\textbf{Frame interpolation.}
Since the neural video field encodes a video as a temporally continuous signal \cite{NeRV, NVP}, we can seamlessly extend the learned signals along the temporal axis, \textit{i.e.}, frame interpolation, without any additional operations. We validate its ability of frame interpolation in \cref{tab:quan} with the UVG-HD dataset \cite{UVG}. Three well-known objective metrics are used for evaluation, including PSNR, SSIM \cite{ssim}, and LPIPS \cite{LPIPS}. We evenly select 300 frames from 600 frames every other frame and calculate average metrics of our interpolated 300 frames with the Ground Truth. Results in the first row of \cref{tab:quan} prove that NVF successfully predicts intermediate frames with competitive accuracy.

\begin{figure}[t]
  \centering
  \begin{subfigure}{0.88\linewidth}
    \includegraphics[width=1.\linewidth]{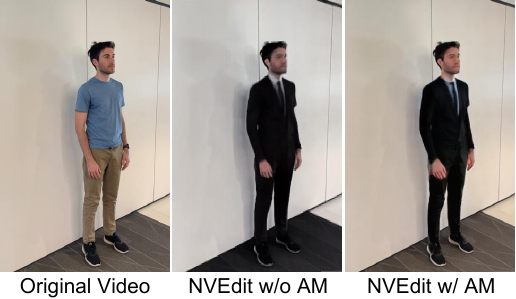}
  \end{subfigure}
  \vspace{-0.6em}
  
  \caption{Results of NVEdit trained with (w/) and without (w/o) our proposed Auxiliary Mask (AM). Without AM, the background is incorrectly changed, while our final version achieves precise local editing.}
  \label{abla}
  \vspace{-1.3em}
\end{figure}

\subsection{Ablation Study}
\label{sec:4.4}
In this section, we conduct ablation research on progressive optimization in the field editing stage, Auxiliary Mask (AM) used in IP2P+, and the adopted T2I model. The first study aims to prove the importance of progressive training in generating temporal consistent videos, and the second demonstrates the potential of our method to improve video editing performance by utilizing new image related technologies. Finally, we replace IP2P+ with another image processing algorithm to validate the substitutability of the T2I model within our framework.

\textbf{Ablation on progressive optimization.}
To investigate the contribution of progressive training, we first fix the random seed of IP2P+ and apply it per-frame, with the results as the control group. Besides, we train an NVEdit but fix the value of $s_P$ in the field editing stage. Results are reported in \cref{tab:abla}, in which we adopt the same objective metrics used in \cref{compare}. One can see that compared to directly using IP2P+, training NVEdit with a fixed $s_P$ actually leads to significant degradation on `Tem-Con', a measure of inter frame consistency. We believe this is because the fixed editing intensity disrupts the prior learned in the video fitting stage. In contrast, as listed in \cref{compare}, the complete NVEdit achieves optimal results across all three metrics compared to these two ablation settings, which proves the effectiveness of our progressive optimization strategy.

\textbf{Ablation on AM.} We have showcased AM and the results of IP2P and IP2P+ in \cref{mask}, here we provide the rendered frames of the same instruction from NVEdit trained w/ and w/o the AM to underscore its importance. Since AM points out the editable regions for local editing, IP2P+ enables more precise editing and produces higher-quality pseudo-GTs to optimize our NVF. Therefore, as depicted in \cref{abla}, NVEdit outperforms the ablation setting, which exhibits a clearer face and an unchanged background. We further provide more visual comparisons between IP2P and IP2P+ in the supplementary material.

\begin{figure}[t]
  \centering
  \begin{subfigure}{1.\linewidth}
    \includegraphics[width=1.\linewidth]{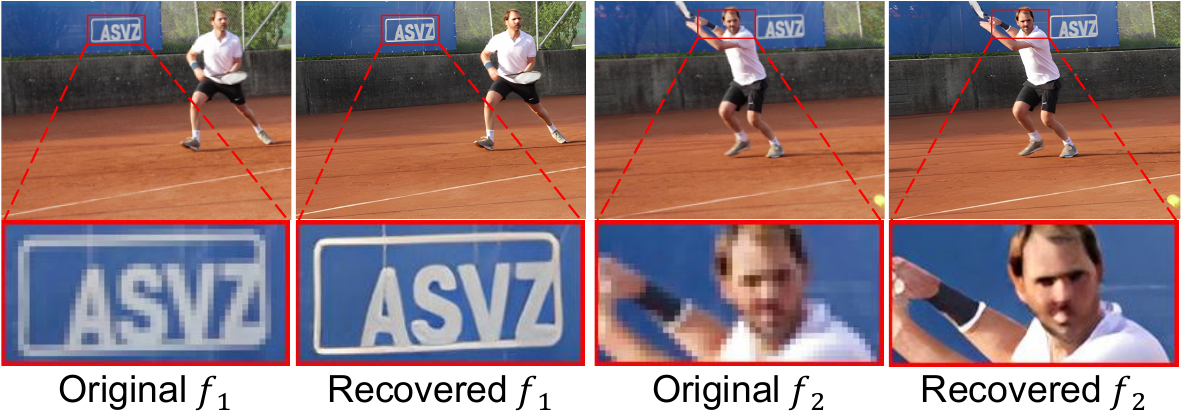}
  \end{subfigure}
  \vspace{-1.8em}
  
  \caption{Video super-resolution results ($\times$4) generated by using an image super-resolution algorithm \cite{realesrgan} to optimize the video field. Two original low-resolution frames and their recovered versions are provided for intuitive illustration.}
  \label{sr}
  \vspace{-1.3em}
\end{figure}

\textbf{Ablation on the T2I model.} Beyond IP2P, various image processing algorithms can be also applied in NVEdit for diverse requirements. Here we take R-ESRGAN \cite{realesrgan} as an example. By utilizing it to conduct image super-resolution on each rendered frame and produce pseudo-GT, our method seamlessly enhances the resolution of videos. As illustrated in \cref{sr}, we select two original frames and provide their recovered versions on the right. To achieve it, in the video fitting stage, NVEdit initially fits the original low-resolution video, then is optimized with the pseudo-high-resolution ($\times$4) GTs in the field editing stage. The final rendered frames are listed on the right for comparison. Besides, we further calculate quantitative results on UVG-HD dataset and report them in the second row of \cref{tab:quan}. Videos are first downsampled by four times and then recovered. These superior results validate the potential of NVEdit that adopting image algorithms to achieve different video tasks in a plug-and-play manner.

\section{Conclusion}
\label{sec:con}
This paper presents a memory-efficient framework \textbf{NVEdit} to edit videos with off-the-shelf image processing methods. We design a neural video field to enable long video editing and learn temporal priors for inter-frame consistency. In order to impart editing effects to the trained video field while preserving its temporal priors, a progressive optimization strategy is developed. Furthermore, to improve editing quality, we upgrade IP2P with the proposed auxiliary mask to benefit its precision in local editing, showcasing remarkable control over editable regions without additional computation, dubbed IP2P+. NVEdit outperforms existing baselines, demonstrating substantial improvements in local editing, temporal consistency, and editable video length. At last, we shed the unique advantages and immense potential of neural representations in video editing, illustrating how they can be synergized with T2I models. Both the video field and IP2P+ can be replaced or improved for various requirements, which inspires future research.

\noindent\textbf{Limitations.} Although we have achieved SOTA consistency in editing, temporal priors are still inevitably affected to some extent in the field editing stage. Moreover, the increased frames require more iterations in the field editing stage, resulting in more time costs. We believe that, in the future, only editing key frames and using frame interpolation of NVF to propagate the editing effect throughout the entire video is a potential acceleration solution.

% Now we have reached the maximum length of an ECCV \ECCVyear{} submission (excluding references).
% References should start immediately after the main text, but can continue past p.\ 14 if needed.

\clearpage  % TODO REVIEW/FINAL: This \clearpage needs to be removed from both review and camera-ready versions.

\bibliographystyle{splncs04}
\bibliography{main}
\end{document}